\documentclass[lettersize,journal]{IEEEtran}
\usepackage{amsmath,amsfonts}
\usepackage{algorithmic}
\usepackage{algorithm}
\usepackage{array}
\usepackage[caption=false,font=normalsize,labelfont=sf,textfont=sf]{subfig}
\usepackage{textcomp}
\usepackage{stfloats}
\usepackage{url}
\usepackage{verbatim}
\usepackage{graphicx}
\usepackage{color,array,amsthm}
\usepackage{orcidlink}
\usepackage{siunitx}
\usepackage{longtable,tabularx}
\usepackage{multirow}
\usepackage{makecell}
\usepackage[noadjust]{cite}

\begin{document}

\title{SSPARE: Space Solar Power Autonomously Reconfigurable Elements}

\author{Dario Tscholl \orcidlink{0000-0002-6286-8185}, Brian C. Gunter \orcidlink{0000-0002-4743-6173}}

\markboth{IEEE Aerospace and Electronic Systems Magazine, April~2024}%
{Shell \MakeLowercase{\textit{et al.}}: A Sample Article Using IEEEtran.cls for IEEE Journals}


\maketitle

\begin{abstract}GEO communication satellites generate significant revenue but can only function reliably for approximately 10 years on orbit. One of the main drivers that limits the reliability of a GEO satellite is the electric power system, and in particular, anomalies related to batteries and degradation of the solar arrays. Given the high cost and relatively short lifespan of GEO satellites, there has been increased research activity towards developing on-orbit servicing systems. However, most of the existing servicing systems are expensive, highly customized, and focus on refueling tasks. On-orbit refueling can be very useful, however, it does not improve satellite reliability which is crucial for long-term missions. Therefore, we propose SSPARE (Space Solar Power Autonomously Reconfigurable Elements), a cost-effective, self-servicing power system. Aside from improving satellite reliability, SSPARE enables to generate up to 6 times more power per launch compared to a traditional GEO communication satellite. This study explores why GEO satellites fail and elaborates on the SSPARE concept. A comparison of SSPARE against a traditional on-orbit servicing mission highlights the benefits of the proposed concept. With humanity striving to become more and more Earth-independent, this work aims to build a foundation for future systems such as large solar power farms on-orbit. \end{abstract}

\begin{IEEEkeywords}
Space Solar Power, Autonomous, Modular, On-Orbit Servicing and Manfucaturing (OSAM), Satellite Power System (SPS), Electronic Power System (EPS), On-Orbit Servicing (OOS), Conceptual Design 
\end{IEEEkeywords}

\section{INTRODUCTION} \label{Introduction}
\begin{figure}[htp]
    \centering
    \includegraphics[width=\columnwidth] {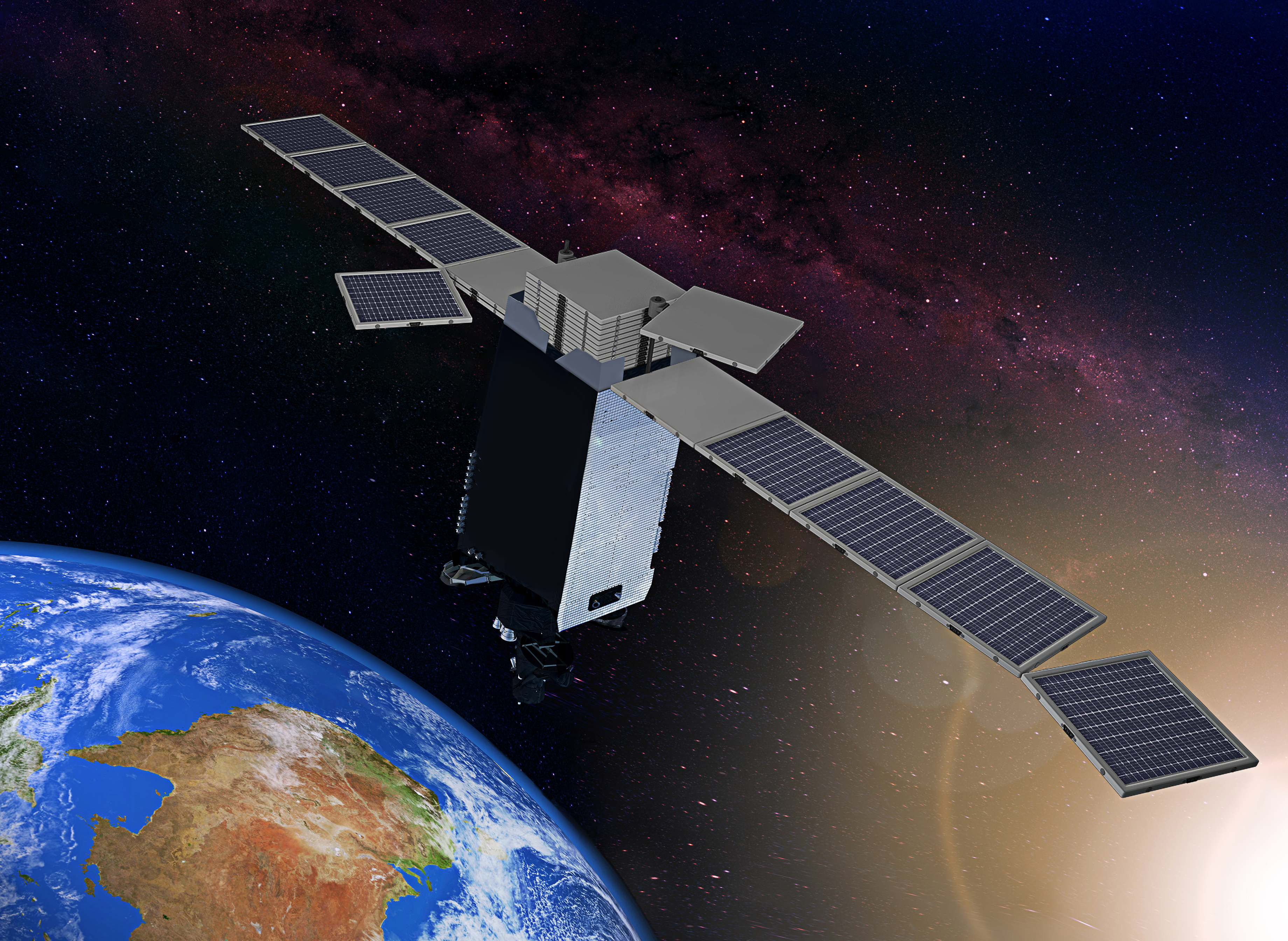} 
    \caption{Illustration of the SSPARE concept}
    \label{System_Example}
\end{figure}

N{\scshape ASA's} current vision to become increasingly Earth-independent and capable of expanding into the solar system is fundamentally driven by the reliability of spacecraft, i.e., have satellites last for long periods with minimal maintenance \cite{nasa_next_2022}. The importance of spacecraft reliability has been widely acknowledged, with numerous prior studies performed concerning satellite system and subsystem reliability \cite{castet_satellite_2009, castet_beyond_2010, dubos_statistical_2010, noauthor_spacecraft_2012}. In \cite{dubos_statistical_2010}, Castet and Saleh found that the reliability of large-satellites steeply decreases after year 7. This sudden drop is also known as wear-out, that appears typical to large satellites. A separate study by the German Aerospace Center (DLR) came to the conclusion that large-satellites won't be able to function reliably for more than 10 years because of limitations in the power system \cite{ellery_case_2008}. Technological advances in battery systems have enabled satellite lifetimes to increase from 7 to 10 years. Extending the life from 10 to 15 years was achieved by using GaAs instead of NiH solar cells, as they are less prone to degradation. \cite{ellery_case_2008}. According to the DLR study, the only viable option to ensure successful long-term missions would be the use of on-orbit servicing (OOS) systems. In recent years, the industry has put significant effort into developing OOS systems, especially for GEO communication satellites. An example for such a GEO servicing mission is Northrop Grumman's MEV-1 \cite{noauthor_companies_2020}. MEV-1 is a servicing satellite with the goal of demonstrating docking in GEO and extending the life of the Intelsat (IS) 901 by 5 more years. Servicing missions like MEV-1 are still very expensive and don't increase the life of a target satellite substantially. Therefore, we propose an innovative satellite power concept called SSPARE (Space Solar Power Autonomously Reconfigurable Elements) that can service itself through the use of stowed spare components, as seen in Figure \ref{System_Example}. Compared to a traditional satellite power system, the power generation and storage happens in one place. By having both the solar panels and battery outside of the satellite, they become easily accessible, enabling autonomous assembly and replacement. Hence, if a power module is broken, SSPARE can service itself and therefore eliminate mission failures due to power anomalies.

The following two subsections further highlight the relevance of the SSPARE concept by going into more detail about GEO satellites and their failure modes. Chapter \ref{Related_Work} elaborates on the current state-of-the-art servicing missions and modular systems. In Chapter \ref{System_Overview} we present a notional concept of our proposed system architecture. Chapter \ref{On-Orbit_Operation} suggests a possible assembly and replacement procedure of SSPARE. In Chapter \ref{Tech_Roadmap}, three major milestones have been identified to guide the development of SSPARE. Lastly, Chapter \ref{Conclusion} points out the significance of our work and its potential on a greater scale.
    
    \subsection{Target Satellites}
    The reason why the industry often focuses on GEO satellites, and primarily communication satellites, is because of the high revenue potential for this satellite class. The 2023 State of the Satellite Industry Report reveals that around 95\% of the \$113 billion in revenue generated by satellite services come from GEO satellites. That means that each GEO satellite can generate millions of dollars in revenue each year. While operation can be lucrative, building a GEO satellite is expensive. The development and launch cost of a GEO satellite is around \$400 million. Thus, the high cost and annual revenue make GEO satellites very attractive for OOS missions. 
    
    \subsection{Cause-Of-Failure}
\begin{figure}[htp]
        \centering
        \includegraphics[width=\columnwidth] {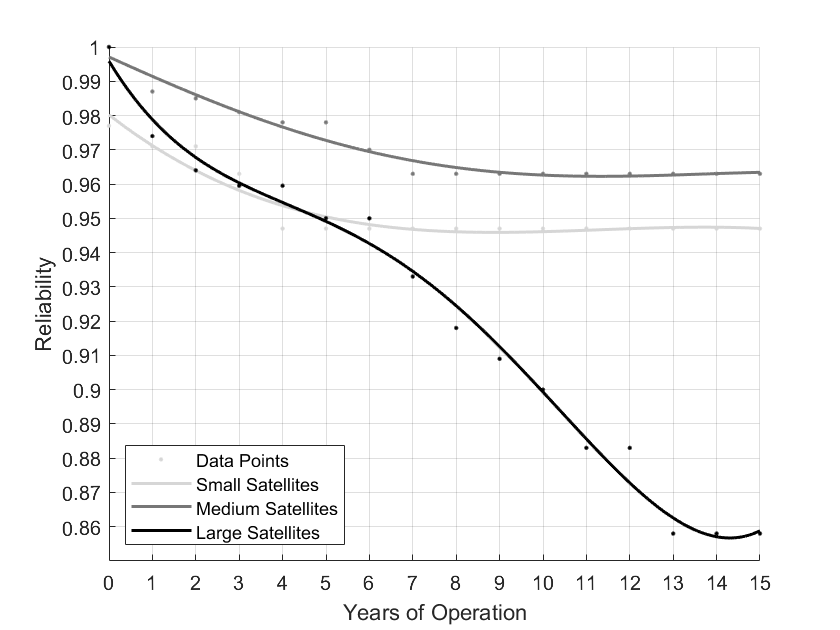} 
        \caption{Reliability of small ($\leq$ \SI{500}{\kilo\gram}), medium (\SI{500}{\kilo\gram}$-$\SI{2,500}{\kilo\gram}) and large ($\geq$ \SI{2,500}{\kilo\gram}) satellites over their years of operation. Clear drop in reliability of large satellites after year 7. Adapted after Dubos et al.}
        \label{Sat_Reliability}
    \end{figure}

    Most openly-available data on satellite failures is on small-satellites. Normally small-satellites are defined as spacecraft weighing anywhere between \SI{1}{\kilo\gram} to \SI{500}{\kilo\gram}. However, some studies about GEO satellite do exist. In the following, we give a qualitative overview of GEO satellite failures by drawing connections between small- and large satellite data. While it's uncertain what percentage of GEO satellites fail, from the data available, it is observed that the reliability of large satellites drops significantly after year 7, as shown in Figure \ref{Sat_Reliability}.

\begin{figure}[htp]
        \centering
        \includegraphics[width=0.9\columnwidth] {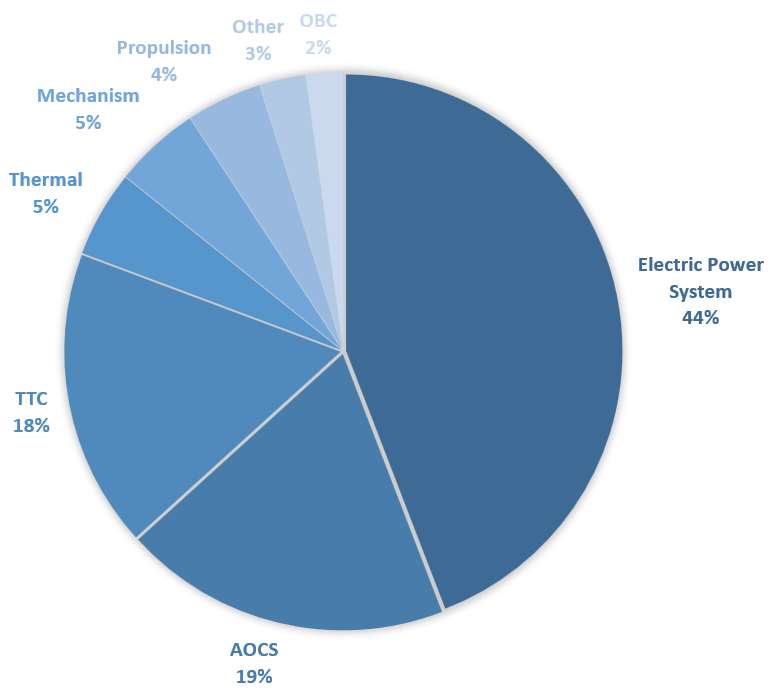} 
        \caption{Satellite failure modes after 10 years of operation. Satellites from 1990-2019 weighing between \SI{40}{\kilo \gram}-\SI{500}{\kilo \gram} were considered. The main failure mode is the electric power system at 44\%. Adapted after Perumal et al.}
        \label{Sat_Failures}
    \end{figure}

    In the case of small-satellites, around 40\% experience partial or complete mission failure \cite{jacklin_small-satellite_2019}. The likelihood of a subsystem experiencing partial or complete mission failure varies over the mission duration. During the first 30 days after launch, the main failure modes are power system (31.5\%), AOCS - attitude and orbit control system (31.7\%), TTC - telemetry, tracking and command (16\%) and mechanisms (15.5\%). After 10 years of operation, the electric power system is clearly the dominant failure mode at 44.1\%. A full breakdown of subsystems that contribute to long-term satellite failure can be seen in Figure \ref{Sat_Failures} \cite{perumal_small_nodate}.

    Insurance claims reveal that the percentage of power system failures is even higher for GEO satellites. According to insurance data, 49\% of all claims come from the power system; with solar arrays contributing 38\% and batteries 11\% \cite{frost_commercial_2004}. Interestingly, those 49\% of power failures make up 62\% of the total insurance cost. It should be noted that this statistic had been minorly skewed by a systemic design error in the solar arrays on the Boeing BSS-702 bus as the platform failed on almost every mission during its first few years of operation. Nonetheless, even when excluding those data points, the power system remains the dominant failure mode in GEO satellites.
    
    Because of the high number of on-orbit failures attributed to power system anomalies, NASA Glenn conducted a detailed study on the causes behind power-related satellite failures \cite{landis_causes_2006}. NASA Glenn divided the power failures into nine different categories and calculated the estimated failure cost generated by each category during 1990-2005. Accumulating the cost of all power failures yields a total loss of around 4.4 billion dollars. Although those numbers apply to small-satellites, Castet and Saleh compared how LEO power system failures compare to GEO power system failures \cite{noauthor_spacecraft_2012}. They showed that for LEO satellites, the power system component failures are quite evenly distributed. However, for GEO satellites, solar array operation (SAO) failure is clearly the dominant failure mode at 69\%, followed by power distribution and battery failure, respectively. Moreover, Castet and Saleh found that GEO satellites generally encounter around 8 times more power failures than LEO satellites. In summary, the lifetime of GEO satellites is limited by the power system and their reliability is strongly driven by SAO failures. 
    
\section{RELATED WORK} \label{Related_Work}
The beginning of on-orbit servicing systems can be found in early Space Shuttle missions in the form of robotic arms. Those systems, like the Shuttle Remote Manipulator System (SRMS), or better known as Canadarm1, date all the way back to the 1980s. Back then, the SRMS was already able to deploy and recover payloads, help construct the International Space Station (ISS) and support astronaut extra vehicular activities (EVAs) \cite{ma_advances_2023}. Over time, robotic OOS arms have greatly matured in terms of perception, dexterity and precision. Current OOS arms like the GITAI S1 are capable of performing general-purpose tasks that allow the assembly of structures and panels, and the operation of switches and cables in space \cite{noauthor_GITAI_2023}. Furthermore, engineers from NASA and Androidnaya Technika have developed entire humanoid OOS robots \cite{noauthor_What_2022, noauthor_FEDOR_2020} that can realize human-robot interactions and operate hardware inside a spacecraft. Aside from these stationary OOS systems, many companies like Motiv Space System, NASA and Northrop Grumman are developing free-floating servicing satellites. Those satellites are usually equipped with one or two robotic arms to repair, maintain, and upgrade a target satellite \cite{noauthor_Motive_2022, noauthor_OSAM-1_2022, noauthor_Robotic_2016}. In 2007, NASA together with DARPA, Boeing and Ball Aerospace demonstrated autonomous rendezvous, capture, refueling and battery transfer between two spacecraft in LEO with the help of a 6-DOF robotic arm \cite{friend_orbital_2008}. Between 2011 and 2020, NASA went a step further and exemplified on-orbit refueling of satellites that were not initially designed for refueling \cite{nasa_robotic_2023}. In 2020 and 2021, Northrop Grumman successfully demonstrated commercial on-orbit refueling in GEO as part of their MEV-1 \& 2 missions. As a next step, Northrop Grumman plans to demonstrate full commercial robotic servicing by 2025 with their mission robotic vehicle (MRV) \cite{noauthor_Robotic_2016, noauthor_Mission_2021}. While servicing missions enjoy a lot of attention, the literature on self-servicing concepts is sparse. Self-servicing concepts that do exist again use the same idea of having a robotic arm to enable self-manipulation \cite{sun_concept_2006}. 

An alternative to repairing a malfunctioning component is to replace it entirely. Replacing something requires a modular system. Modular, and self-reconfigurable systems in particular, are widely considered to be one of the grand challenges in robotics. Modular robots can be classified into different architectures based on their geometric arrangement \cite{yim_modular_2007}. In this work we focus primarily on a chain architecture since our power modules will likely be connected in a string topology. Most architectures, including the chain architecture, use their elements to reconfigure themselves. To do so, there are different ways of locomotion such as sliding horizontally \cite{byoung_kwon_an_em-cube_2008, chiang_modular_2001}, vertically \cite{hosokawa_self-organizing_1998} or both \cite{murata_m-tran_2002}, by pivoting \cite{gipe_reconfigurable_2022} or through deformation \cite{suh_telecubes_2002}. Over the years, researchers focused mainly on pivoting systems because sliding or deforming system do not scale well. Recent research studies have developed cube-shaped pivoting robots that are very light-weight \cite{nisser_electrovoxel_2022}, have passive-locking capabilities \cite{oroke_technology_2023} and can even jump \cite{romanishin_3d_2015}, all while being able to move and reconfigure themselves in three dimensions. Modular robots are generally versatile, robust and low-cost which makes them attractive for space applications \cite{yim_modular_2003, noauthor_polymorphic_2014}. For the past twenty years, the main space application and focus has been on increasing the aperture size of space telescopes. To solve that problem, several modular space telescope concepts have been proposed but none of them have been successfully demonstrated yet \cite{basu_proposed_2003, miller_assembly_2008, lee_architecture_2016, underwood_autonomous_nodate}.

In recent years, the interest in modularity in space has shifted from telescopes to space system interfaces. A modular interface would enable a building block approach to assemble modules in space. Standardized, reliable \cite{diaz-carrasco_diaz_sirom_2023} and lightweight \cite{kortmann_building_2015} modular interfaces have been developed up to technology readiness level (TRL) 7. On a subsystem level, researchers have also demonstrated, that it is possible to decouple satellite subsystems (i.e. solar module, battery module and payload module) and connect them modularly \cite{lim_modular_2018}.

\begin{figure}[hb]
    \raggedleft
    \includegraphics[width=0.8\columnwidth] {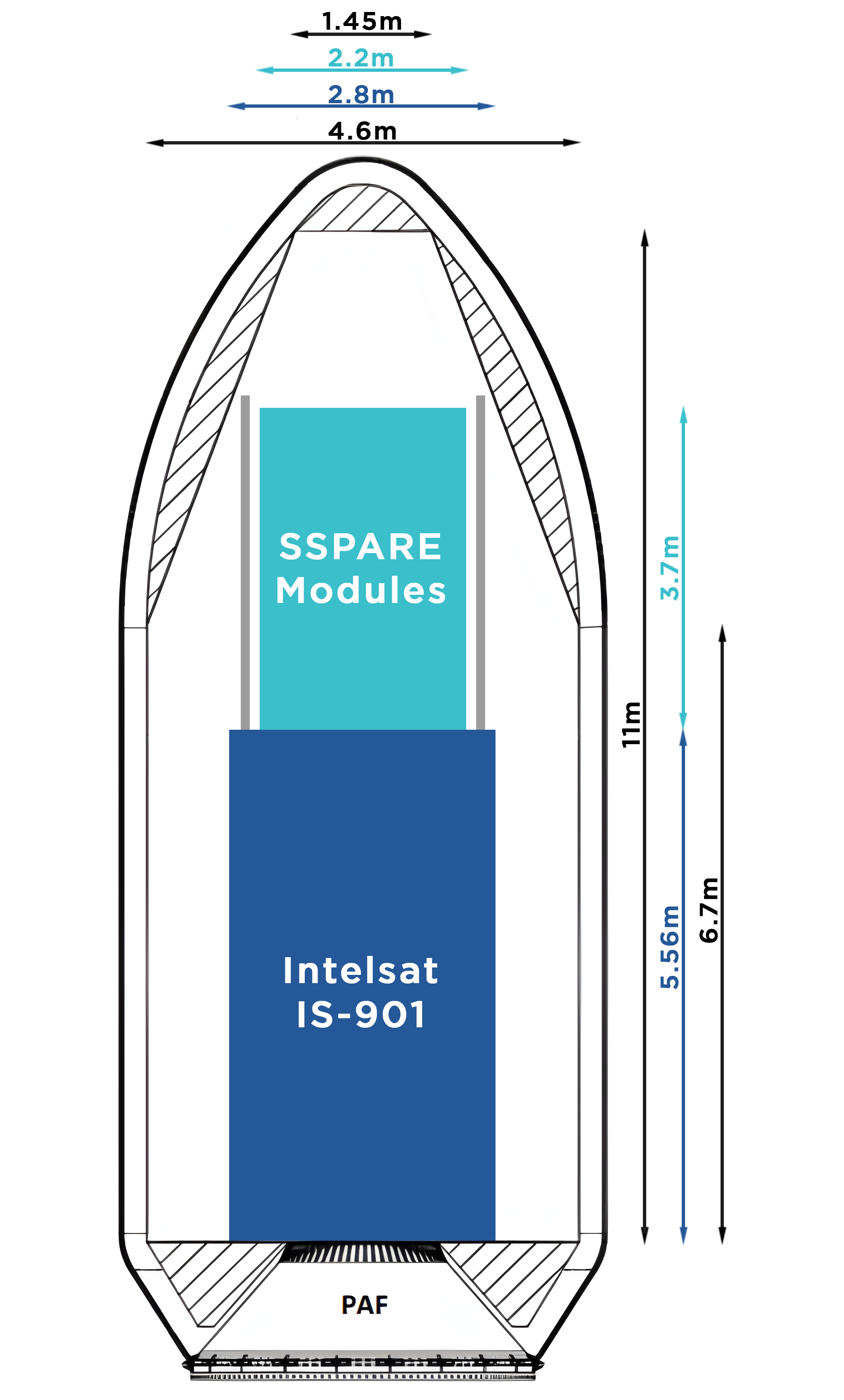} 
    \caption{Intelsat 901 with SSPARE modules stored inside of a standard Falcon fairing. The grey columns on the sides of the SSPARE modules represent the rods. The rods are part of the unloading system and limit the stacking height. The rods reach a tapered fairing diameter of \SI{2.6}{\meter}.}
    \label{falcon-fairing}
\end{figure}

\section{SYSTEM OVERVIEW} \label{System_Overview}
As introduced in Section \ref{Introduction}, our goal is to develop a modular, self-assembling power system to increase long-term satellite reliability. To achieve this goal, we propose SSPARE. The SSPARE concept can be divided into two parts: 1) The solar power modules and 2) a system to stow and unload the solar power modules. In the following, we elaborate on each of the two parts and propose a preliminary sizing of the system. In the end, we compare a satellite with the SSPARE architecture against a traditional satellite being serviced. Because on-orbit servicing of GEO satellites has only been demonstrated on the Intelsat 901 and 10-02, the preliminary sizing of the SSPARE modules has been done based on the IS-901.

    \subsection{Space Solar Power Modules}
    A traditional electric power system (EPS) consists of solar panels, regulators and a battery system. Components like batteries and regulators are inherently fixed to a satellite's structure which makes accessing them almost impossible from the outside. For example, if the batteries malfunction, the whole satellite will fail. Thus, more solar panels won't make a traditional satellite more resilient. Therefore, we suggest combining the solar panels and the power storage/distribution into a single module, referred to as a space solar power module (SSPM). Space solar power modules possess the key ability to move independently to their intended slot once they are deployed. This allows for autonomous assembly and replacement of entire solar panel arrays. Moreover, the modular nature of SSPARE enables to exploit the full payload capacity during launch by stacking the modules on top of the satellite, as seen in Figure \ref{falcon-fairing}.

    Depending on the packing efficiency of the modules, this could lead to requiring a smaller launch vehicle and thus, driving down financial and environmental cost. Intelsat satellites like the upcoming Galaxy 13R/Horizons-4, for example, only make use of around 45\% of the Falcon 9's available payload volume, and around 42\% of the Falcon 9's available mass to orbit capacity. Maximizing the payload space and mass would provide the same mission with more power capability and/or redundancy for no additional launch costs.
    
    Another benefit of SSPARE's modularity is that if a module breaks during operation, the SSPMs can replace themselves immediately. This presents a significant advantage over external servicing missions, which can result in lengthy delays of weeks or even years depending on the availability of a servicer. Furthermore, SSPARE's architecture enables seamless resupply of additional solar power modules, which is highly beneficial should all spare elements be exhausted. Theoretically, this resupply capability offers the potential to extend the amount of power generated almost indefinitely.

    \subsection{Unloading System} 
    There are many potential ways to unload the space solar power modules. Possible implementations include various types of a robotic mechanism, a dispenser, free-flyer or tether-based concepts. As mentioned in Chapter \ref{Related_Work}, several free-flying concepts have been proposed in the past but have proven to be extremely challenging to implement. A dispenser concept lacks natural redundancy, meaning that if something blocks the dispenser, the whole system stops working. Although a two-sided tether-based concept would be more reliable, it adds unnecessary complexity to the system. Furthermore, tether-based concepts are difficult to scale-up. 

\begin{figure}[htp]
        \raggedleft
        \includegraphics[width=\columnwidth] {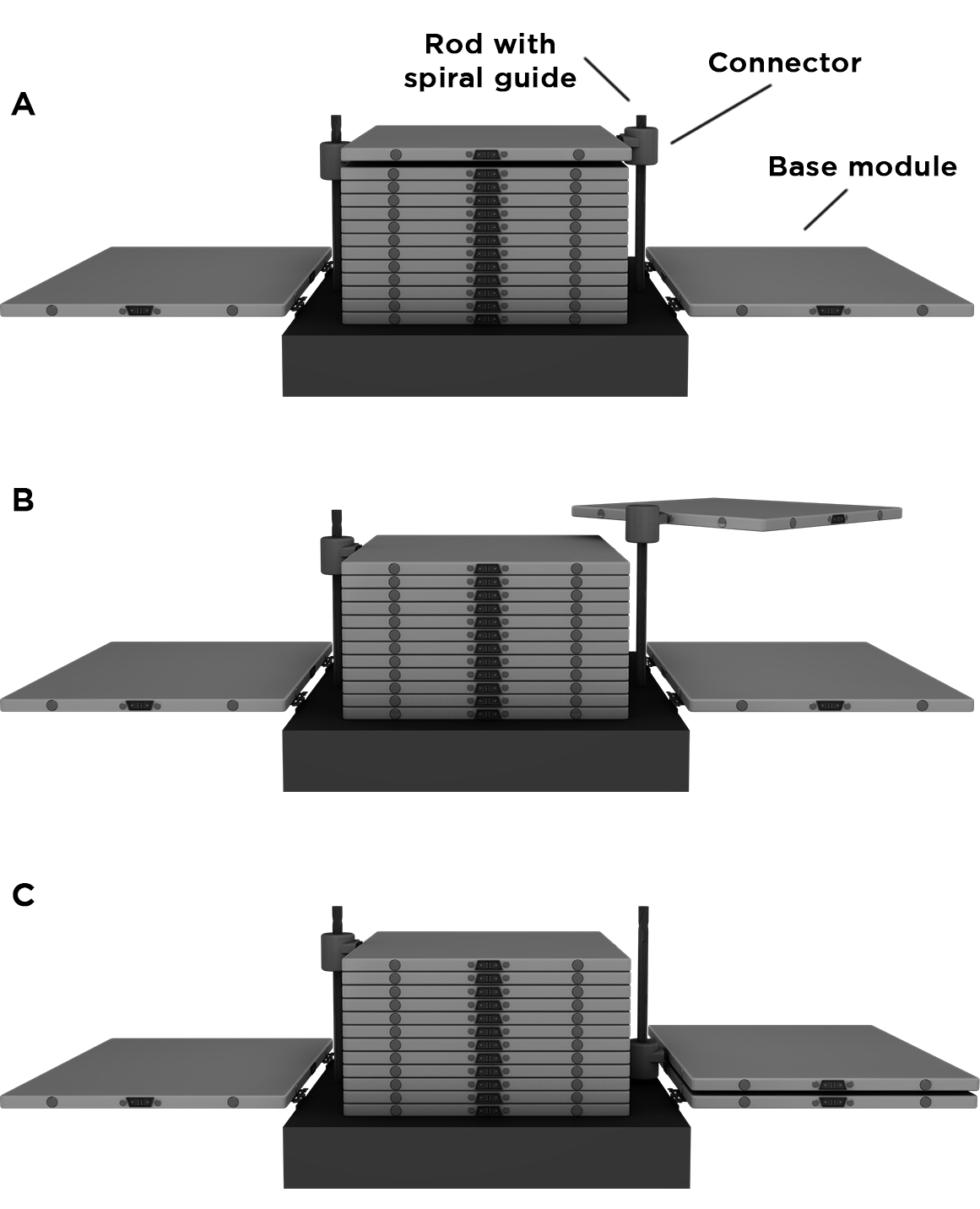} 
        \caption{Proposed unloading system of SSPARE. A) The space solar power modules are stacked on top of the satellite. The right connector docks to the top module and lifts it up. B) The connector moves up and down the rod. The spiral guide at the top  of the rod enables the connector to turn 180 degrees. C) The connector places the module on the base module from which it can unfold and move into place.}
        \label{Unload_Mechanism}
    \end{figure}

    Therefore we propose the use of a robotic mechanism. More precisely, a one degree of freedom forklift-inspired system that offers redundancy and low system complexity. For the forklift-inspired system we assume the solar power modules to be stacked and initially stored on top of the satellite. The stack is separated by two rods opposite to each other. Each rod has a connector; a linear actuator that can move up and down. The connector utilizes the same docking interface as the power modules use to connect to each other during assembly and operation. To make sure the elements are kept together in orbit, at least one connector is always attached to the top element of the stack, exerting a small down-force. Alternatively, other mechanical connections like frangibolts can also be used to secure the elements before use. To transport the modules, the connector moves to the top of the rod where a spiral guide enables the connector to turn \SI{180}{\degree} without the need of an additional actuator. Turning \SI{180}{\degree} is necessary to move the modules from the center to the outside of the satellite. A close-up of the described system is shown in Figure \ref{Unload_Mechanism}.

    On its way down, the connector puts the element on a base module that connects the solar power modules with the main body of the satellite. The two base modules, one on each side, are initially folded down to the side of the satellite during launch. A base module has two functions. Internally, it contains non-critical electronic components to manage and distribute power to the other subsystems. Externally, it serves as a base from which the elements can begin to unfold and move into their designated place. 
    
    \subsection{Preliminary Sizing} \label{Preliminary_Sizing}
    A preliminary sizing of the SSPARE modules yields a square shape with a length, width and height of 2.2/2.2/0.09 \SI{}{\meter}, respectively. From a thermal perspective, the small height of the modules is advantageous as it prevents a large thermal gradient between the top and bottom surface of the modules. This is particularly important for the batteries which require a specific operating temperature range. Measures to keep the batteries within their desired temperature range include radiators at the bottom of the modules and insulation, like multi-layer insulation (MLI). The square shape was selected to allow pivoting around the edges and building entire two-dimensional SSPM arrays. The dimensions of the SSPM were primarily driven by the size of the Intelsat 901. The IS-901 has a length, width and height of 2.8/3.5/5.56 \SI{}{\meter}, respectively. With a solar panel surface of \SI{4.41}{\meter \squared}, a solar irradiance of \SI{1361}{\watt \per \meter \squared} and an assumed conversion efficiency of 0.25, each module is capable of generating around \SI{1.5}{\kilo \watt}. Thus, 6 solar power modules would be enough to provide a power output equivalent to the traditional IS-901. Because the SSPMs can be stacked on top of the satellite, 37 modules can fit inside a standard Falcon fairing, considering each module takes up \SI{0.1}{\meter} of vertical space. An arrangement of the IS-901 with SSPARE in a standard Falcon fairing is shown in Figure \ref{falcon-fairing}. All 37 solar power modules together hold the capacity to generate up to \SI{55.5}{\kilo \watt}. In only a single launch, that is six times the amount a traditional IS-901 can produce.

    \subsection{Cost-Benefit Analysis} \label{CostBenefit_Analysis}
    While the SSPARE concept has many advantages, it adds additional cost and mass to a mission. An initial cost evaluation of the solar power modules reveals a total of \$450,000 per module, with the predominant cost attributed to the solar cells. Covering the top surface of an SSPM requires around 1,350 cells, amounting to approximately \$400,000. The estimated cost for the primary and secondary structure is around \$40,000, while the batteries and electronics are assumed to be around \$10,000. 
    In terms of mass, each module is anticipated to weigh roughly \SI{125}{\kilo \gram}, where the structure and batteries constitute the majority of the mass. Given the volume of a SSPM, the mass of the primary and secondary structure is approximately \SI{108.58}{\kilo \gram}. The estimated mass for the batteries is around \SI{13.3}{\kilo \gram}. The SSPARE concept is not an extension but a redesign of the power system. Therefore, only the unloading system and the spare modules add additional cost and mass to a mission, as the necessary hardware just gets moved from the inside to the outside of the satellite. 
    
    Assuming we equip an IS-901 with 2 base modules and 10 SSPMs, where four of them are spares, the SSPARE concept would add an additional 1.9 million dollars and \SI{750}{\kilo \gram} of mass to the mission (without considering the unloading system). Given that an Intelsat 901 costs around \$400 million and that the satellite can generate an annual revenue of tens of millions, the cost of SSPARE is relatively small. When comparing SSPARE to MEV-1, the cost-effectiveness of SSPARE becomes clear. Northrop Grumman's MEV-1 servicing vehicle costs Intelsat 13 million dollars annually, totaling \$65 million for the full mission duration of five years. Because MEV-1 is a demonstration mission, those \$65 million are only a fraction of the actual cost, as Northrop Grumman is covering most of the mission's expenses.

    Although SSPARE and MEV-1 have different servicing objectives, the mission of both concepts is the same; to extend the life of a target satellite. MEV-1 is costly, deployed during the later stages of a satellite's operation and offers only a marginal extension to the satellite's operational years. SSPARE on the other hand is cost-effective and specifically designed to mitigate the primary failure mode of GEO satellites, right from the start of its operation. While SSPARE is not primarily meant for performing trajectory corrections, simulations in NASA's General Mission Analysis Tool (GMAT) confirmed that SSPARE modules could also be configured to enable station keeping, if equipped with electric propulsion or gas thrusters. Table \ref{OSS_overview} sums up this section by providing a comparison of different system characteristics between three IS-901 configurations.

    \begin{table*}[h]
        \begin{tabular}{ |m{5.1cm}||m{3.2cm}|m{4cm}|m{3.2cm}| } 
             \hline
             \textbf{System Characteristics} & \textbf{Traditional Intelsat} & \textbf{Intelsat with MEV-1 Servicer} & \textbf{Intelsat with SSPARE} \\
             \Xhline{1.2pt}
             Main body dimensions (m/m/m) & 2.8 / 3.5 / 5.6 & 2.8 / 3.5 / 5.6 & 2.8 / 3.5 / 6.6 \\
             \hline
             Maximum power (kW) & 8.6 & 8.6 & 15 \\
             \hline
             Development \& launch cost (\$MM) & 400 & 465 & 402 \\
             \hline
             Launch mass (kg) & 4,725 & 7,051 & 5,475 \\ 
             \hline
             Life expectancy (years) & 17 & 22 & 30+ \\
             \hline
        \end{tabular} \\
    \caption{\label{OSS_overview} Comparing different system characteristics of a traditional Intelsat 901 with its servicing mission (MEV-1) against a potential SSPARE configuration. This particular SSPARE configuration contains 10 space solar power modules, where 4 of them are spares.}
    \end{table*}

    Lastly, it's not uncommon that GEO satellites cost much more than \$400 million. In a recent article about a potential merger, Intelsat and SES are considering to develop a \$10 billion satellite \cite{noauthor_ses_2023}. In such a case, SSPARE would enable long-term satellite reliability at very little extra cost.

\section{ON-ORBIT OPERATION} \label{On-Orbit_Operation}
Just like there are different ways of unloading the modules, there are various options of assembling and replacing them. As seen in Section \ref{Related_Work}, pivoting is the most versatile, robust and scalable solution in modular robotics. However, other solutions like magnetic levitation (maglev) or wheeled locomotion are possible. There are different types of maglev, namely electromagnetic suspension (EMS), electrodynamic suspension (EDS), inductrack and diamagnetic levitation. Neither inductrack nor EDS are able to levitate objects at a standstill. Therefore, wheels would be required to accelerate which increases the system complexity. Aside from the fact that EMS systems rely on active electronic stabilization to remain stable, their magnet design wraps around the rail which takes away the versatility aspect from modular systems. Finally, diamagnetic levitation allows an object to levitate freely given the magnetic field is strong enough. Technology that makes use of this material property, like Diamagnetic Micro Manipulation (DM3) \cite{pelrine_diamagnetically_2012} for example, has so far been limited to micro-scale robots. Although NASA JPL plans to scale this concept to meso- and eventually meter scale robots \cite{hsu_2d-compliant_2022}, this technology is still in its early stages. One of the biggest challenges of this technology is the payload capacity. In the absence of gravity, however, this constraint vanishes. Therefore, diamagnetic levitation could conceivably be a viable solution in the future. 

    \begin{figure}[htp]
        \raggedleft
        \includegraphics[width=\columnwidth] {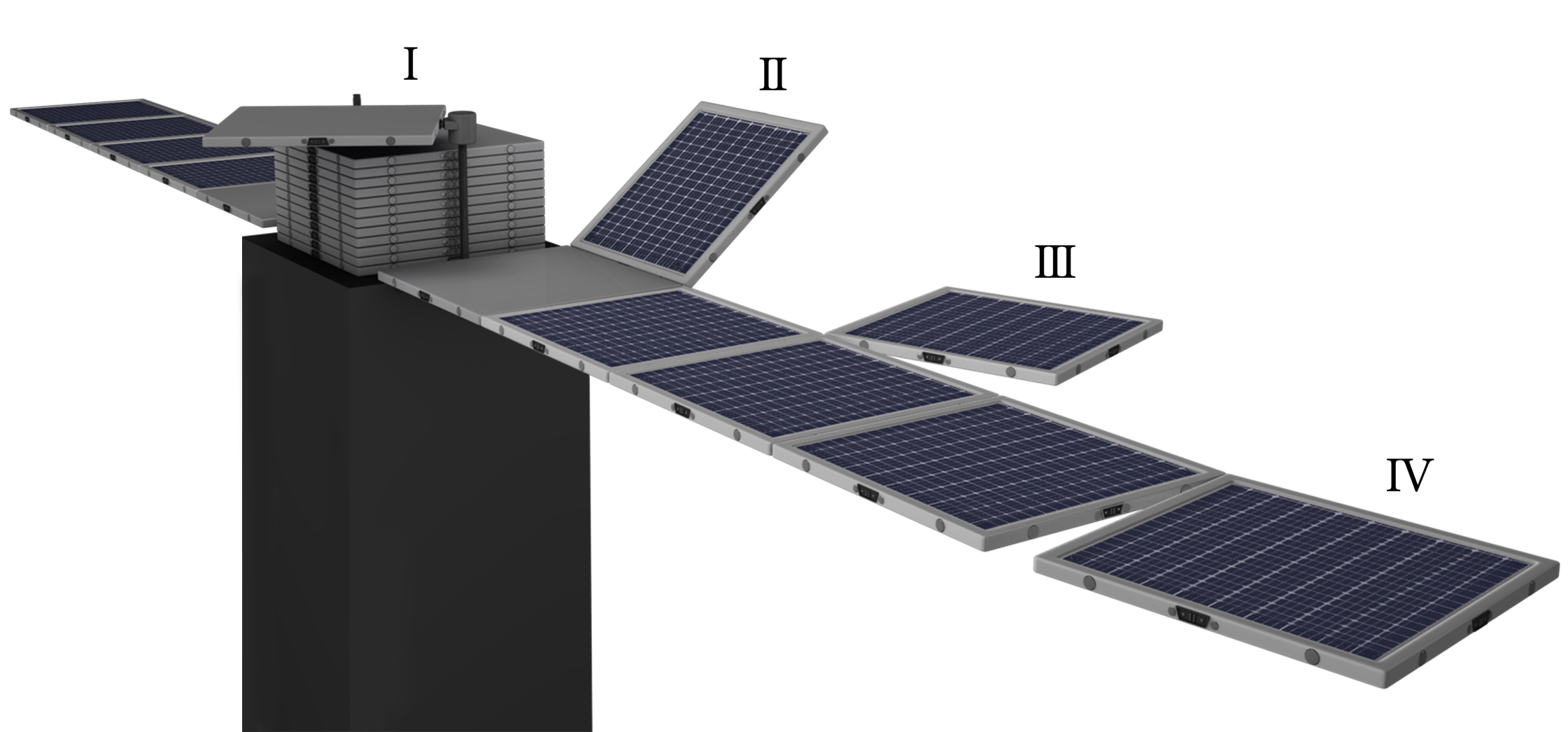} 
        \caption{Assembly of space solar power modules using pivoting. I) A module is being deployed from the stack and is put on the base module. II) The Module unfolds and with that starts to pivot into position. III) The Module pivots around the other SSPMs. IV) The Module connects to the last module in line to form typical chain architecture.}
        \label{Assembly_Proceedure}
    \end{figure}

    \subsection{Assembly Procedure} 
    As already mentioned in Section \ref{System_Overview}, the space solar power modules will be stacked on top of the satellite. From there, a forklift inspired system will move one element at a time from the stack onto one of the base modules. Once the element is on a base module, it will start to pivot around that module and the other SSPMs to form a chain architecture. Once the element is in place, the connector on the rod moves back to facing the stack and securing the next element in line. From there, that same process repeats on the opposite side. The process keeps alternating between each side until the desired amount of elements has been placed.

    \subsection{Replacement Procedure} \label{Replacement}
    One of the key features of SSPARE is its ability to service itself when a module is broken. In this context, broken means that either the module has stopped responding or the module has suffered physical damage (e.g. array or battery failure). To monitor the health of the modules, each SSPM will contain various sensors (e.g. temperature, IMU, distance) and be subject to an overarching sensor node design. While the individual modules communicate with each other over wire or Bluetooth, a central heartbeat protocol will be used to monitor communication across all modules. 
    
\begin{figure}[htp]
    \raggedleft
    \includegraphics[width=\columnwidth] {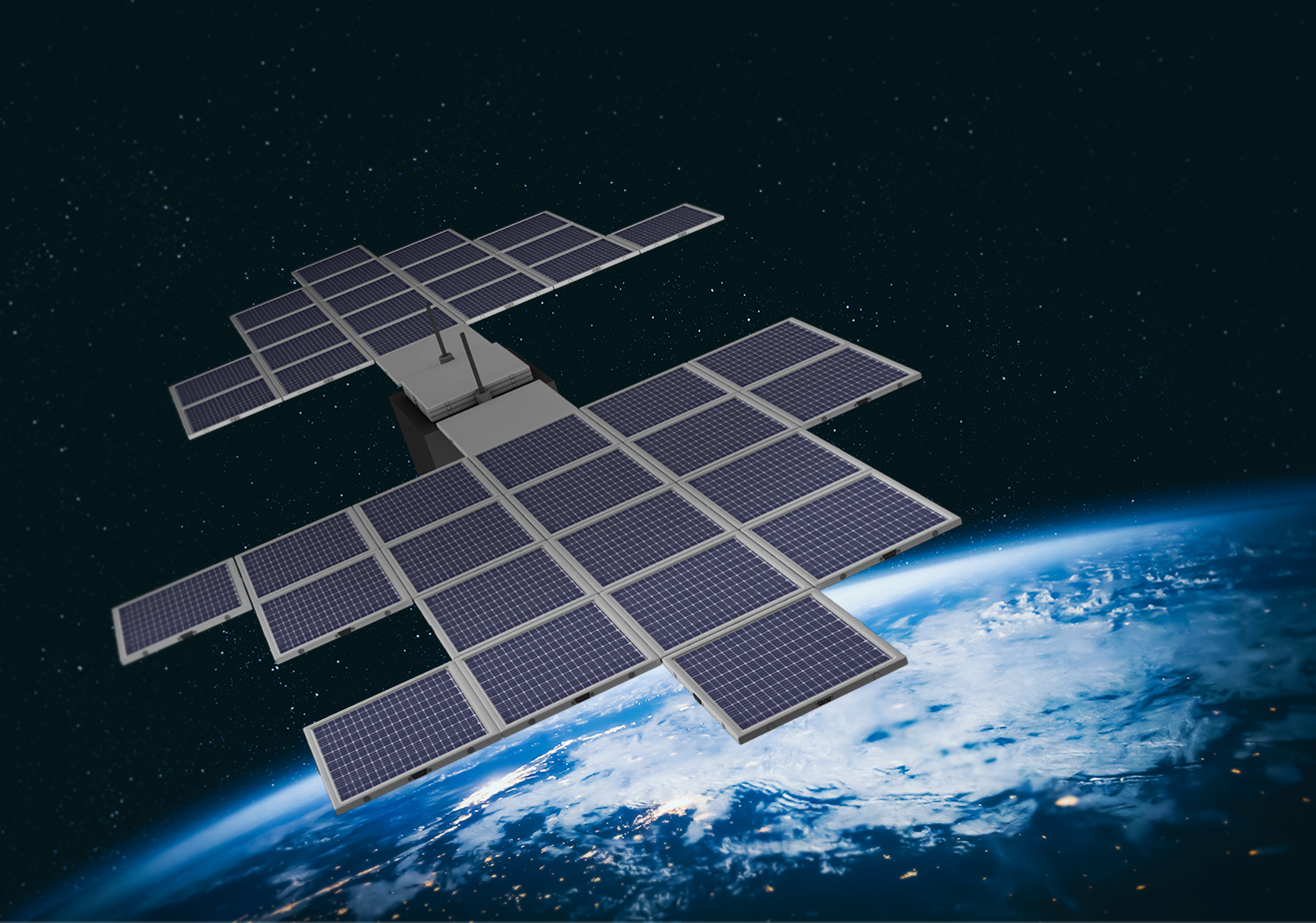} 
    \caption{SSPARE mockup with the majority of its modules deployed. In this configuration, 38 SSPMs are generating 57 kW of power; 6 times more power than a traditional Intelsat.}
    \label{SSPARE_Full}
\end{figure}

    If a module breaks, instead of swapping out the broken module, a new module will be attached to one of the sides of the broken one. To get there, the new module will follow the same assembly procedure as described above. If the broken module is not responding, the newly attached module can trigger a sequence that enables the current to flow through the broken module. A possible implementation includes the use of a cascaded H-bridge converter. A Cascaded H-bridge converter is a multilevel voltage converter that offers redundancy as an individual H-bridge can be bypassed without impacting the overall quality of the voltage. Because SSPMs can be connected to each other's sides, SSPARE can also deviate from the typical string topology. For example, SSPMs can take on any arbitrary pattern as illustrated in \ref{SSPARE_Full}.

\section{TECHNOLOGY ROADMAP} \label{Tech_Roadmap}
The SSPARE concept can be broken down into five major building blocks: Locomotion, power system, autonomy, unloading system and the structural and thermal design of the modules. While certain blocks like the unloading system are already well developed in one form or another, other blocks like the locomotion and power system need more attention. Although researchers have built modular robots in the past, the development of the space solar power modules will face unique challenges. Therefore, we have identified three key milestones that will drive the development of the modules:

\begin{enumerate}
    \item Building a small-scale prototype to iterate on an optimal locomotion strategy.
    \item Developing a modular power system that can generate, store and transfer power across several space solar power modules
    \item Demonstrating meter-scale autonomous locomotion of the space solar power modules in zero-gravity given SSPARE's flat-disk shaped proportions
\end{enumerate}

To reach those three milestones, a prototype to determine an optimal locomotion strategy will have to be build. A technical demonstrator, including a modular power system and ideally sized to a 12U cubesat, will allow for a potential small-scale full-system test on orbit or onboard the ISS. After successfully demonstrating the concept in cm-scale, the concept can then be scaled-up to meter-scale.

\section{CONCLUSION} \label{Conclusion}
Building and launching GEO satellites is expensive, however, operation can be very lucrative. Due to the substantial development cost and annual profits, the goal is to maximize the operational lifetime of GEO satellites. A closer look at at the lifetime of GEO satellites has revealed that they experience a sudden decline in reliability after operating on-orbit for 7 years. The reason for that stems from the electronic power system and specifically from anomalies with solar array operations. It turns out that power system anomalies are responsible for almost two thirds of all insurance claims and therefore, the power system in not only the most common but also the most expensive failure mode. To solve this issue, we propose SSPARE (Space Solar Power Autonomously Reconfigurable Elements). SSPARE stands out from other servicing systems by being self-servicing, cost-effective and designed to enhance the lifetime proactively rather than reactively. In contrast to traditional servicing systems that focus on extending operation towards the end of a satellite's life, a system equipped with SSPARE boosts its reliability right from the start. Aside from eliminating mission failures due to power system anomalies, SSPARE's modular nature comes with several other advantages. SSPARE is able to make full use of a launcher's payload capacity, allowing a satellite to generate significantly more power. Thus, more power can lead to greater mission assurance or increased science return. On orbit, SSPARE has the flexibility to adjust the number of modules in use based on the mission's power demand. Because SSPARE can assemble modules in any desired configuration, SSPARE even has the potential to build entire space solar power farms.

Between 1995-1997, NASA actually conducted a detailed study on space solar power farms. The goal of that study was to investigate the economic and environmental feasibility of building a space solar power farm and integrating the power generated into terrestrial power grids \cite{mankins_fresh_1997}. NASA came to the conclusion that large space solar power structures would be environmentally feasible but not economically viable. The main drivers behind the economic viability were the cost of accessing space and the launch rate associated with it. Since then, companies such as Rocket Lab and SpaceX have driven down the mass to orbit cost substantially. In fact, If SpaceX's Starship program succeeds, the cost of accessing space would be low and the launch rate high enough to overcome the economic limitations that held back the development in the 1990s. With that, our space solar power modules would be a key building block to build such structures. Moreover, many companies like the Air Force Research Laboratory (AFRL) are actively working on technology to beam large amounts of power from space to Earth \cite{air_space_2023}. Those efforts in combination with SSPARE have the potential to change the way of how we harness and distribute space solar power in the future. 

\bibliographystyle{IEEEtran}
\bibliography{main_manuscript}
\end{document}